\documentclass[11pt,twocolumn,twoside]{article}
\usepackage{fully3d}
\usepackage{amsmath,amsfonts}
\usepackage{bm}

\begin{document}

\title{Geometric Constraints Enable Self-Supervised Sinogram\\ Inpainting in Sparse-View Tomography}

\author[1]{Fabian~Wagner}
\author[1]{Mareike~Thies}
\author[1]{Noah~Maul}
\author[1]{Laura~Pfaff}
\author[2]{Oliver~Aust}
\author[3]{Sabrina~Pechmann}
\author[4]{Christopher~Syben}
\author[1]{Andreas~Maier}

\affil[1]{Pattern Recognition Lab, Friedrich-Alexander-Universität Erlangen-Nürnberg, Germany}

\affil[2]{Department of Rheumatology and Immunology, Friedrich-Alexander-Universität Erlangen-Nürnberg, Germany}

\affil[3]{Fraunhofer Institute for Ceramic Technologies and Systems IKTS, Forchheim, Germany}

\affil[4]{Fraunhofer Development Center X-Ray Technology EZRT, Erlangen, Germany}

\maketitle
\thispagestyle{fancy}


\begin{customabstract}
The diagnostic quality of computed tomography (CT) scans is usually restricted by the induced patient dose, scan speed, and image quality. Sparse-angle tomographic scans reduce radiation exposure and accelerate data acquisition, but suffer from image artifacts and noise. Existing image processing algorithms can restore CT reconstruction quality but often require large training data sets or can not be used for truncated objects. This work presents a self-supervised projection inpainting method that allows optimizing missing projective views via gradient-based optimization. By reconstructing independent stacks of projection data, a self-supervised loss is calculated in the CT image domain and used to directly optimize projection image intensities to match the missing tomographic views constrained by the projection geometry. Our experiments on real X-ray microscope (XRM) tomographic mouse tibia bone scans show that our method improves reconstructions by $3.1$--$7.4\,\%$/$7.7$--$17.6\,\%$ in terms of PSNR/SSIM with respect to the interpolation baseline. Our approach is applicable as a flexible self-supervised projection inpainting tool for tomographic applications.
\end{customabstract}


\section{Introduction}
\label{sec:introduction}

Computed tomography (CT) scanners allow the reconstruction of unknown 3D object density distributions from a set of acquired X-ray projection images. Most clinical applications use filtered back projection (FBP)-based reconstruction algorithms that require a dense angular sampling to meet the precondition of the analytical algorithm. Although decreasing the number of measured projection images can be beneficial to reduce patient dose, improve acquisition speed, and reduce motion effects \cite{wagner2022monte}, it introduces noise and image artifacts that can impair diagnostic value.\\
Different image processing algorithms were proposed to restore the image quality of scans acquired with reduced angular sampling and dose, intervening at different stages of the CT reconstruction pipeline. A first set of algorithms operates directly on the acquired sinogram data with the goal to upsample the number of projection images \cite{wei20202,zang2021intratomo}. Consistency conditions on CT projection data were applied to limited-angle acquisitions to improve image quality while preserving consistency with the measured data \cite{huang2017restoration,aichert2015epipolar}. However, such methods are often insensitive to in-plane artifacts and only work for non-truncated data. A second group of methods apply pure image post-processing algorithms to improve the overall quality of noisy reconstructions acquired with reduced dose \cite{wagner2022ultra,zhang2021noise2context,jeon2022mm}. To make CT reconstruction pipelines compatible with gradient-based data-driven training, differentiable FBP operators were presented that allow propagating a loss calculated on the reconstructed image back to the raw sinogram data \cite{syben2019pyro,thies2022calibration}. Such known operators \cite{maier2019learning} allow training CT pipelines employing neural networks on the sinogram and the reconstruction simultaneously \cite{wagner2022benefit}. Existing approaches use neural networks or other denoising operators to restore image quality \cite{thies2022learned,huang2019data} or make use of conventional inpainting techniques to increase the angular projection sampling artificially \cite{li2012strategy}.\\
So far most learning-based models are trained supervisedly on large-scale low and high-quality paired data sets \cite{wei20202}. However, multiple self-supervised training approaches exist that circumvent the need for paired training data. Noise2Noise \cite{lehtinen2018noise2noise} calculates a loss metric from two independent noisy image representations. Noise2Inverse \cite{hendriksen2020noise2inverse} and other works \cite{wu2021low} extend this principle to tomographic CT settings by splitting projection data into two independent sets. The resulting reconstructions are regarded as image representations with independent noise realizations and allow for deriving a self-supervised denoising loss. Only a few self-supervised inpainting techniques exist and even fewer are applied to CT problems \cite{zang2021intratomo,kim2022streak}.\\
In this work, we present a self-supervised CT projection inpainting method. We regard missing projection images as trainable data tensors that are updated to be consistent with the measured data. Reconstructing them with a differentiable FBP operator allows deriving a self-supervised loss in the image domain with the real measured sparse-angle reconstruction. Backpropagating that loss through our fully differentiable reconstruction pipeline allows directly optimizing the missing projection data tensor to generate additional projection views. Our contributions are the following:
\begin{itemize}
  \item We present a self-supervised projection inpainting method based on a differentiable FBP pipeline.
  \item We propose to directly optimize missing CT projection information through the fixed projective geometry of a differentiable FBP operator.
  \item We evaluate our method on high-resolution cone-beam X-ray microscope (XRM) ex-vivo mouse tibia bone data.
\end{itemize}

\begin{figure*}[t]
  \centering
  \includegraphics[width=1.0\textwidth]{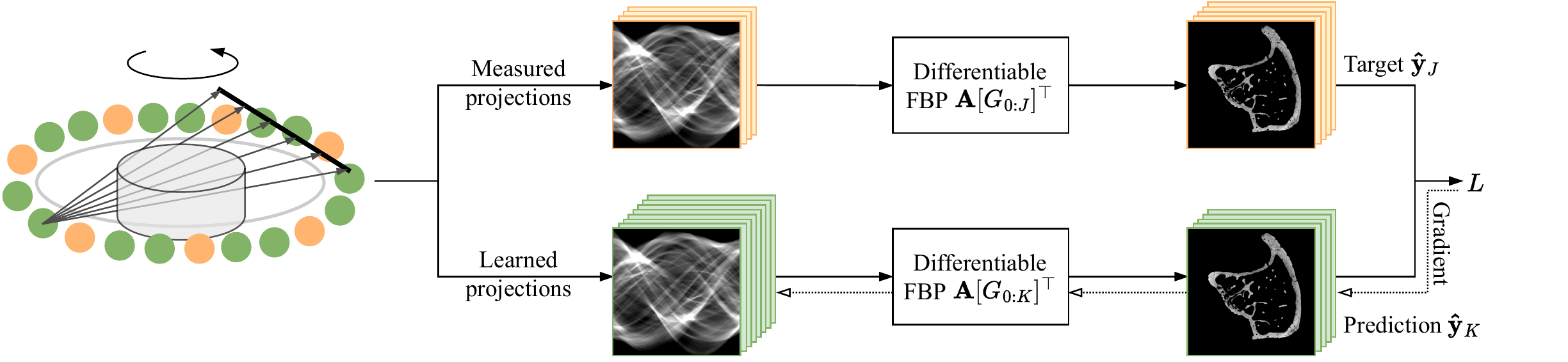}
  \caption{Illustration of the proposed self-supervised projection inpainting scheme. A target image is reconstructed from a sparsely sampled CT scan, e.g., one-third of the usual number of projections from a high-resolution scan (orange data). The intensities of intermediate missing projections are regarded as trainable parameters (green projections). Further, a self-supervised loss is calculated between the two reconstructed projection stacks and backpropagated through a differentiable FBP operator to the inpainted projections to update them consistently with the measured data.}
  \label{fig:pipeline}
\end{figure*}

\section{Methods}
\label{sec:methods}

\subsection{Differentiable Reconstruction Pipeline}
\label{ssec:reco}
The acquisition of CT projection images $\mathbf{p}_j$ with $j \in \{0, \dots, J\}$ can be written as
\begin{align}
    \mathbf{p}_j = \mathbf{A}[G_j] \mathbf{y} + \mathbf{n}_j
\end{align}
with the true scanned object $\mathbf{y}$, noise vector $\mathbf{n}_j$, and the forward operator $\mathbf{A}$ conditioned to the projection geometry $G$. The inverse CT problem aims to reconstruct the unknown object from the acquired set of $J$ projections using the adjoint system matrix $\mathbf{A}^\top$ and the filter operator $\mathbf{K}$
\begin{align}
    \mathbf{\hat{y}}_J = \frac{1}{J} \sum_{j} \mathbf{A}[G_j]^\top \mathbf{K} \mathbf{p}_j\enspace.
\end{align}
Using a differentiable backprojection operator \cite{syben2019pyro} allows backpropagating a loss metric $L$ derived on the reconstructed image $\mathbf{\hat{y}}$ to the raw CT projection data using the chain rule
\begin{align}
    \frac{\partial L}{\partial \mathbf{p}_j} = \frac{\partial L}{\partial \mathbf{\hat{y}}} \frac{\partial \mathbf{\hat{y}}}{\partial \mathbf{p}_j}\enspace.
\end{align}
Hence, models working in the projection domain or the projection data itself can be directly modified using gradient-based optimization.

\subsection{Optimizing Consistent Sinogram Information}
\label{ssec:methods}
Following the Noise2Inverse \cite{hendriksen2020noise2inverse} pipeline, a self-supervised loss $L$ can be derived from the reconstructions of two independent stacks of CT projection images $\mathbf{p}_j$ ($j \in \{0, \dots, J\}$) and $\mathbf{p}_k$ ($k \in \{0, \dots, K\}$). In the reconstruction domain, only their image content but not their noise realization is correlated which allows deriving a loss between both reconstructed images to assess the consistency of the image contents. Other works \cite{hendriksen2020noise2inverse} minimize that loss to train the weights of denoising models $f(\cdot, w)$ in the projection and the reconstruction domain self-supervisedly
\begin{align}
    \underset{w}{\text{argmin }} L \left(\mathbf{\hat{y}}_J, \mathbf{\hat{y}}_K\right)\enspace.
\end{align}
In the sparse-angle tomographic setting where only a limited number of $J$ projection images was measured, we propose to minimize the following expression to optimize the missing set of $K$ noise-free projection images $\mathbf{p}_k$
\begin{align}
    \underset{\mathbf{p}_k}{\text{argmin }} L \left(\mathbf{\hat{y}}_J, \mathbf{\hat{y}}_K\right) = \underset{\mathbf{p}_k}{\text{argmin }} L \left(\mathbf{\hat{y}}_J[G_{0:J}], \mathbf{\hat{y}}_K[G_{0:K}]\right)\enspace.
\end{align}
We regard the pixel intensities of the inpainted projection images $\mathbf{p}_k$ as trainable weights and optimize them in a self-supervised and data-driven way as illustrated in Fig.\,\ref{fig:pipeline}. To enable backpropagating the gradient to the projection images, we use a differentiable FBP operator \cite{syben2019pyro} within our XRM reconstruction pipeline \cite{thies2022calibration}. Due to the well-defined projection geometries $G_j$ and $G_k$ for the individual projection images $\mathbf{p}_j$ and $\mathbf{p}_k$, the proposed pipeline is constrained to optimize missing angular projection data in between the measured sparse-angle projections to predict realistic reconstructions $\mathbf{\hat{y}}_K$ close to $\mathbf{\hat{y}}_J$. Although the pipeline can in theory converge to an identity solution where inpainted projections equal the true measured data, this solution is not favored by the derived pixel-wise loss as both independent reconstructions would be slightly misregistered through the different sets of view geometries $G_{0:J}$ and $G_{0:K}$.

\section{Experiments}
\label{sec:experiments}

\subsection{Data}
\label{ssec:data}
The used data set consists of five high-resolution X-ray microscope (XRM) cone-beam scans of ex-vivo mouse tibia bones. Investigating bone structures on the micrometer scale is instructive for understanding the cause and progression of bone-related diseases on the cellular level as well as for developing adapted therapies. Lacunar bone structures, visible in Fig.\,\ref{fig:predictions} as tiny holes in the bone, are in particular of interest as they contain osteocyte cells that are heavily involved in the bone-remodeling process \cite{gruneboom2019next}. To resolve Lacunae reasonably well, $1401$ projection images are acquired, which accumulates to a total acquisition time of around $14\,\text{h}$. Here, neither acquisition time nor induced sample dose allow for desired in-vivo investigations \cite{wagner2022monte}. Sparse-angle CT acquisitions in combination with self-supervised projection inpainting algorithms can enable faster low-dose acquisition while preserving a high reconstruction quality. The used XRM scans image the tibia bone of mice close to the knee joint and contain truncated information of the proximal fibula bone in some projective views.\\
The study was performed in line with the principles of the Declaration of Helsinki. Approval was granted by the Ethics Committee of FAU Erlangen-Nürnberg (license TS-10/2017).

\subsection{Training}
\label{ssec:training}
We evaluate the effectiveness of our method on five XRM mouse tibia scans using half ($50\,\%$ dose) and one-third ($33\,\%$ dose) of the available $1401$ projections for all compared inpainting strategies. Projection images are reconstructed using the public cone-beam XRM reconstruction framework of Thies~et~al.~\cite{thies2022calibration}. Nearest neighbor interpolation and trilinear interpolation operators are taken from the PyTorch framework and compared to our proposed self-supervised projection inpainting method. We initialize the missing projection intensities with interpolated projections to accelerate the optimization and start from a reasonable reconstruction. Subsequently, projections $\mathbf{p}_k$ are registered in the PyTorch graph as trainable parameters and are updated using a stochastic gradient descent optimizer with learning rate $0.1$ without momentum. Further, we use the mean absolute error as loss function $L$. Quantitative performance metrics are calculated with respect to the high-resolution XRM images reconstructed from all available projections.

\section{Results and Discussion}
\label{sec:results}

\begin{table}[b]
  \centering
  \begin{tabular}{lcc}
  \toprule
  $50\,\%$ dose & PSNR & SSIM \\
  \midrule
  Nearest neigh. int. & $27.0 \pm 0.4$ & $0.620 \pm 0.008$ \\
  Trilinear int. & $27.8 \pm 0.4$ & $0.656 \pm 0.008$ \\
  Optimized projections & $\bm{29.0 \pm 0.5}$ & $\bm{0.729 \pm 0.009}$ \\
  \bottomrule
  \addlinespace[2ex]
  \toprule
  $33\,\%$ dose & PSNR & SSIM \\
  \midrule
  Nearest neigh. int. & $25.8 \pm 0.5$ & $0.571 \pm 0.011$ \\
  Trilinear int. & $26.2 \pm 0.5$ & $0.585 \pm 0.011$ \\
  Optimized projections & $\bm{26.6 \pm 0.5}$ & $\bm{0.615 \pm 0.013}$ \\
  \bottomrule
  \end{tabular}
  \caption{Quantitative reconstruction results (mean $\pm$ std) calculated between the prediction starting from one half and one third of the full number of projections and the high-resolution reconstruction that is regarded as ground truth.}
  \label{tab:xrm_results}
\end{table}\noindent
We compute the commonly used quantitative image quality metrics peak signal-to-noise ratio (PSNR) and structural similarity index measure (SSIM) to evaluate reconstructions with respect to the high-resolution ground truth images reconstructed from all available projections. Mean $\pm$ standard deviation of the used five XRM scans is provided in Tab.\,\ref{tab:xrm_results} for both investigated sparse-angle settings. The calculated metrics indicate improved reconstruction quality of the optimized XRM projections across both sparse-angle tomographic settings and over the compared reference methods. In our experiments, optimized projection images improve the PSNR by $3.1$--$7.4\,\%$ and the SSIM by $7.7$--$17.6\,\%$ relative to the nearest neighbor interpolation baseline.\\
\begin{figure}[t]
  \centering
  \includegraphics[width=1.0\columnwidth]{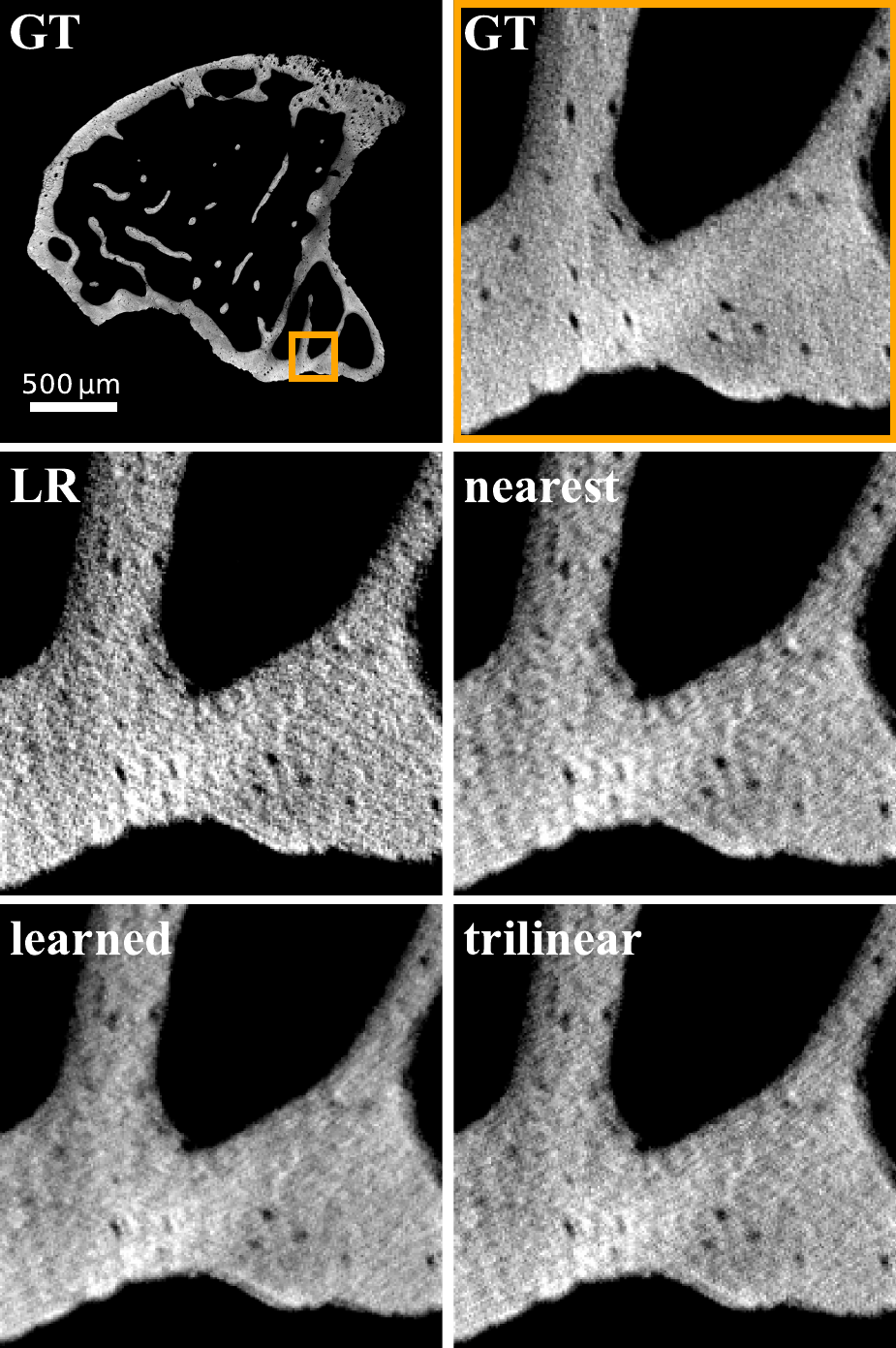}
  \caption{Reconstructions of mouse tibia bone XRM scans. The magnified region of interest is highlighted in the overview slice. The ground truth (GT) and low-resolution (LR) images are calculated from all and one-third of the available projection images respectively. Nearest, trilinear, and optimized denote the different reconstructions from the interpolated and optimized CT projection images.}
  \label{fig:predictions}
\end{figure}
Qualitative results are compared on the reconstructions of the inpainted projection images in magnified regions of interest in Fig.\,\ref{fig:predictions}. Biologically interesting lacunar structures are visible as small dark holes within the bone tissue. Whereas nearest neighbor interpolation only slightly improved the noise level within the bone region, trilinear interpolated and optimized projections further reduce reconstruction noise. Although the bone reconstructed from the optimized projections appears a bit smoother compared to trilinear interpolation, only small improvements are visible.\\
None of the compared methods contain deep neural networks or require any form of pre-training on additional data, reference data, or a learned prior. In our proposed method, projection intensities are directly changed using gradient-based optimization within the fully differentiable reconstruction pipeline. In contrast to other techniques employing view consistency, we believe that our self-supervised pipeline is not limited to circular CT trajectories but can be applied to more difficult acquisition geometries as long as there is a differentiable reconstruction operator at hand. In addition, three of the five used bone scans contain truncated residual parts of the proximal fibula bone which are only visible in some projection views which would severely disturb most projection consistency-based algorithms. Our experiments show that the present data truncation can be handled well by our proposed projection optimization method. Further work is required to fully evaluate the generalizability of end-to-end optimization of CT projections and compare it to existing projection consistency-based methods or deep learning-based models.

\section{Conclusion}
\label{sec:conclusion}

In this work, we present a truly self-supervised projection inpainting technique to improve the reconstruction quality of sparsely acquired CT projection data. Our method allows directly optimizing projection image intensities through a differentiable FBP operator and a self-supervised loss metric calculated from two independently reconstructed projection stacks. 
The proposed pipeline requires further evaluation to prove its clinical effectiveness. Nevertheless, it has great potential over existing self-supervised algorithms as additional regularization can be applied and data consistency is enforced.

\section{Acknowledgments}
\label{sec:acknowledgments}

This work was supported by the European Research Council (ERC Grant No. 810316) and a GPU donation through the NVIDIA Hardware Grant Program.


\printbibliography

\end{document}